\documentclass{llncs}

\newcommand{\A}{} 
\newcommand{\BT}{} 
\newcommand{\Ed}{} 
\newcommand{\JT}{} 
\newcommand{\VY}{} 
\newcommand{\Bem}[1]{} 
\newcommand{\IN}{in:} 

\begin{document}
 
%
%
%
\title{Transferrable Plausibility Model - A Probabilistic Interpretation of Mathematical Theory of Evidence}
%
%
\author{   Mieczys{\l}aw Alojzy K{\l}opotek}
\institute  {Institute of Computer Science, Polish Academy of Sciences,   Warszawa, Poland,\\
\email{klopotek@ipipan.waw.pl}
}

%
%
%

\maketitle              

\begin{abstract}
This paper suggests a new interpretation of the Dempster-Shafer theory in terms of probabilistic interpretation of plausibility.
A new rule of combination of independent evidence is shown and its preservation of interpretation is demonstrated. %
\footnote{
This is a preliminary version of the paper:
\\
M.A. K³opotek: Transferable Plausibility Model - A Probabilistic Interpretation of Mathematical Theory of Evidence O.Hryniewicz, J. Kacprzyk, J.Koronacki, S.Wierzchoñ: Issues in Intelligent Systems Paradigms Akademicka Oficyna Wydawnicza EXIT, Warszawa 2005 ISBN 83-87674-90-7, pp.107--118 
}
\end{abstract}

\section{Introduction}

Dempster Rule of Independent Evidence Combination has been criticized for its failure to conform to probabilistic interpretation ascribed to belief and plausibility function.
Among those verifying DST (Dempster-Shafer-Theory, \cite{Dempster:67,Shafer:76}) critically were 
Kyburg \cite{Kyburg:87},
Fagin \cite{Fagin:89}, 
Halpern \cite{Halpern:92},
Pearl \cite{Pearl:90},
Provan \cite{Provan:90},
Cano \cite{Cano:93},
just to mention a few. 

As a way out of those difficulties, we proposed in a recent book co-authored by S.T.Wierzchon \cite{MKP:Wierzchon:02:mono}  three proposals for an empirical model of DST: 
\begin{itemize}
\item "the marginally correct approximation". 
\item "the qualitative model"
\item "the quantitative model"
\end{itemize}

The marginally correct approximation assumes that the belief function shall constitute lower bounds for frequencies, though only for the marginals, and not for the joint distribution. Then, the reasoning process is expressed in terms of the so-called Cano et al. conditionals - a special class of conditional belief functions that are positive. This approach implies modification of the reasoning mechanism, because the correctness is maintained only by reasoning forward. Depending on the reasoning direction we need different "Markov  trees" for the reasoning engine. 

Note that lower/upper bound interpretations have a long tradition for DST \cite{Dempster:67,Kyburg:87} and have been heavily criticized \cite{Halpern:92}. The one that we presented in our book differs from the known ones significantly as we insist on different reasoning schemes (hypertrees) depending on which are our target variables, whose values are to be inferred. This assures overcoming of the basic difficulties with lower/upper bound interpretations.  

Our qualitative approach is based on the earlier rough set interpretations of DST, but makes a small and  still significant distinction. All computations are carried out in a strictly "relational" way, that is,  indistinguishable objects in a database are merged (no object identities). The behavior under reasoning fits strictly the DST reasoning model. Factors of well established hypergraph representation (due to Shafer and Shenoy \cite{Shenoy:94}) can be expressed by relational tables. Conditional independence is well defined. However, there is no interpretation for conditional belief functions in this model.

Rough set interpretations \cite{Skowron:94} were primarily developed for interpreting the belief function in terms of decision tables. However, the Dempster-rule of evidence combination was valid there only for the "extended decision tables", not easily derived from the original ones. In our interpretation, both the original tables and the resultant tables dealt with when simulating Dempster-rule are conventional decision tables and the process of  combining of decision tables is a natural one (relational join operator).  

Our rough set based interpretation  may be directly applied in the domain of multiple decision tables: independence of decision variables or Shenoy's  conditional independence in the sense of DST may serve as an indication of  possibility of decomposition of the decision table into smaller but   equivalent  tables. 
 
Furthermore, it may  be applied in the area of Cooperative Query Answering \cite{Ras:91}.  The problem there is that  a query posed  to  a local relational database  system may contain an unknown attribute. But, possibly, other co-operating  database systems know it and may explain it to the queried system in terms of known attributes, shared by the various systems. The uncertainties studied in the decision tables arise here in a natural way and our interpretation  may be used to measure these uncertainties in terms of DST (as a diversity of 
support). Furthermore, if several co-operating systems respond, then the queried system may calculate the overall uncertainty measure using DST combination of measures of individual responses.  

The quantitative model assumed that the objects possess multivalued properties which are then lost in some physical properties and these physical processes are described by DST belief functions (see e.g. \cite{MKP:02g}).. 

The quantitative model assumes that during the reasoning process one attaches labels to  objects
hiding some of their properties. There is a full agreement with the reasoning mechanism of DST. Conditional 
independence and conditional belief functions are well defined. We have also elaborated processes that 
can give rise to well-controlled graphoidally structured belief functions. 
Thus, sample generation for DST is possible. 
We elaborated also learning procedures for discovery of graphoidal structures from data. 

The quantitative model seems to be the best fitting model for belief functions created so far. 

This frequency model differs from what was previously considered 
\cite{Smets:92,Smets:94}
in that it assumes that reasoning in DST is connected with updating of variables for individual cases. This is different from e.g. reasoning in probability where reasoning means only selection of cases. In this way, failures of previous approaches could be overcome. 

Many authors \cite{Shafer:90ijar,Smets:92} question the need for an empirical model for DST and point rather to theoretical properties of DST considered within an axiomatic framework seeking parallels with the probability theory.  
Though it is true that the probability theory may be applied within the framework of Kolmogorov axioms and quite useful results are derived in this way, one shall still point out that the applicability of probability theory 
is significantly connected with frequencies. Both frequencies considered as "naive probabilities", ore ones being probabilities "in the limit".   
Statistics is clearly an important part of the probabilistic world.

All three interpretations share a common drawback – they are not sensu stricto probabilistic. 
In the current paper we make an attempt of a purely probabilistic vision of plausibility function. 

\section{Basics of the Dempster-Shafer Theory}

We understand DST measures in a standard way
(see \cite{Shafer:76}). 
Let $\Xi$ be a finite  set of elements called elementary events. 
Any subset of $\Xi$ is a composite event, or hypothesis. $\Xi$ be called also 
 the 
frame of discernment.

\begin{definition} \label{bpadef}\label{DEFm} \cite{Shafer:76}
Let $\Omega$ be a finite  set of elements called elementary events. The set $\Omega$ is called frame of discernment.
Any subset of $\Omega $ be a composite event.  \\
A basic probability assignment (bpa) function is any function m:$2^\Omega
\rightarrow [0, 1]$ such that  $$  \sum_{A \in 2^\Omega } m(A)=ONE   
\qquad
  m(\emptyset)=0, \qquad 
\forall_{A \in 2^\Omega} \quad  0 \leq  \sum_{A \subseteq B} m(B)$$
We say that a bpa             is vacuous iff $m(\Omega)=ONE$ and $m(A)=0$ for every 
 $A\ne\Omega$. 
\end{definition}
      
If ONE is equal 1, then we say that the belief function is normalized, otherwise not (but ONE must be positive). 

\begin{definition}  \cite{Shafer:76}
Let a belief function be defined as Bel:$2^\Omega \rightarrow [0,1]$ so that 
 $Bel(A) = \sum_{B \subseteq A} m(B)$.
Let a plausibility function be Pl:$2^\Omega \rightarrow [ 0,1]$  with 
$\forall_{A \in 2^\Omega} \  Pl(A) = ONE-Bel(\Omega-A )$, a
 commonality function be Q:$2^\Omega-\{\emptyset\} \rightarrow [0,1]$ with 
 $\forall_{A \in 2^\Omega-\{\emptyset\}} \quad Q(A) = \sum_{A \subseteq B}
m(B)$. 
\end{definition}

\begin{definition}  \cite{Shafer:76}
The           Rule of Combination of two Independent Belief Functions 
$Bel_{E_1}$,
 $Bel_{E_2}$ Over the Same Frame of Discernment (the so-called Dempster-Rule), 
 denoted 
    $$Bel_{E_1,E_2}=Bel_{E_1} \oplus Bel_{E_2}$$ 
 is defined as follows: :
$$m_{E_1,E_2}(A)=c \cdot  \sum_{B,C; A= B \cap C} m_{E_1}(B) \cdot 
m_{E_2}(C)$$ (c - constant normalizing the sum of $m$ to 1). 
\end{definition}

Under multivariate settings $\Xi$ is a set of vectors in n-dimensional space 
spanned by the set of variables {\bf X}=\{ $X_1, X_2, \dots X_n$\}. 
If $A\subseteq\Xi$, then by projection  $A^{\downarrow {\bf Y}}$ 
of the set  $A$ onto a subspace spanned 
by the set of variables ${\bf Y}\subseteq {\bf X}$ we 
understand the set $B$ of vectors from $A$ projected onto {\bf Y}.   
Then marginalization operator of DST is defined as follows: \linebreak
$m ^{\downarrow {\bf Y} }(B)= 
\sum_{A; B=A  ^{\downarrow {\tiny
X} }} m(A)$.

\begin{definition} \label{ShaferCond}
 (See \cite{Shafer:90b}) Let B be a subset of $\Xi$, called 
evidence,
 $m_B$ be a basic probability assignment such that $m_B(B)=1$ and $m_B(A)=0$
for any A different from B. Then the conditional belief function $Bel(.||B)$
representing the belief function $Bel$ conditioned on evidence  B 
is defined
as: $Bel(.||B)=Bel \oplus Bel_B$. 
\end{definition}

\section{New Rule of Evidence Combination}

Let us suggest now a totally new approach to understanding belief functions.

We assume the following interpretation of the plausibility function:
$Pl_{\xi}(A)$ is the maximum probability that an element from the set of events $A$ occurs, given the evidence $\xi$, where we assume the apriorical probability of all elementary events is equal. 
Let ${{\xi}1}$ and ${{\xi}2}$ be two independent bodies of evidence, which are represented numerically by plausibility functions
$Pl_{{\xi}1}$ and $Pl_{{\xi}2}$ over some frame of discourse $\Omega$.
We would like to obtain such an evidence updating rule $\oplus_{Pl}$ that 
$Pl_3=Pl_{{\xi}1}\oplus_{Pl}Pl_{{\xi}2}$ would have the semantics that under that interpretation 
$Pl_3(A)$ is the maximum probability that an element from the set of events $A$ occurs, given the evidence $Pl_1$, $Pl_2$ under the least conflicting evidence.

Let us study in detail this assumption.
First of all we have to tell what we mean by independent evidence.
Let $\omega$ be an elementary event from the frame of discernment $\Omega$.  
The body of evidence ${\xi}1$ is independent of the body ${\xi}2$ if, for each $\omega \in \Omega$, the probability of occurrence of evidence ${\xi}1$ is independent of the occurrence of evidence ${\xi}2$. 
So we say that $Pr({\xi}1 \land {\xi}2 | \omega)=Pr({\xi}1 | \omega) \cdot Pr( {\xi}2 | \omega)$. 

How shall we understand the evidence, however.
For any $A\subseteq \Omega)$ should hold $Pl_{\xi}(A)\ge Pr(A|\xi)$. 
Consequently, by the way, $Pl_{\xi}(A)+Pl_{\xi}(\Omega/A)\ge 1$.

Now observe that $Pr(\omega_1 \lor \omega_2|\xi)=Pr(\omega_1 |\xi)+Pr( \omega_2|\xi)$. 
As a consequence, we have always that 
$Pl_{\xi}(\{\omega_1\})+Pl_{\xi}(\{\omega_2\}) \ge Pl_{\xi}(\{\omega_1,\omega_2 \})$. 

Let us now turn to combining independent evidence.
$$Pr(\omega | {\xi}1 \land  {\xi}2)= %
Pr({\xi}1 \land  {\xi}2 | \omega) \cdot \frac{Pr(\omega)}{Pr( {\xi}1 \land  {\xi}2)} %
Pr({\xi}1|\omega)\cdot Pr({\xi}2 | \omega) \cdot \frac{Pr(\omega)}{Pr( {\xi}1 \land  {\xi}2)} %
$$ $$ %
Pr(\omega|{\xi}1)\cdot Pr(\omega|{\xi}2) \cdot \frac{Pr( {\xi}1) \cdot Pr({\xi}2)}{Pr( {\xi}1 \land  {\xi}2)\cdot Pr(\omega)} %
$$
So we can conclude that 
$Pl_{{\xi}1\land{\xi}2}(\omega)=Pl_{{\xi}1}(\omega) \cdot Pl_{{\xi}2}(\omega) \cdot c$ where $c$ is a normalizing factor (which needs to be chosen carefully).

But what about $Pr(\omega_1 \lor \omega_2 | {\xi}1 \land  {\xi}2)$ ? 
We know that 
$Pr(\omega_1 \lor \omega_2 | {\xi}1 \land  {\xi}2) %
= Pr(\omega_1  | {\xi}1 \land  {\xi}2) %
+ Pr( \omega_2 | {\xi}1 \land  {\xi}2) %
$ hence 
$$Pr(\omega_1 \lor \omega_2 | {\xi}1 \land  {\xi}2) %
$$ $$ %
 Pr(\omega_1|{\xi}1)\cdot Pr(\omega_1|{\xi}2) \cdot \frac{Pr( {\xi}1) \cdot Pr({\xi}2)}{Pr( {\xi}1 \land  {\xi}2)\cdot Pr(\omega_1)} %
+Pr(\omega_2|{\xi}1)\cdot Pr(\omega_2|{\xi}2) \cdot \frac{Pr( {\xi}1) \cdot Pr({\xi}2)}{Pr( {\xi}1 \land  {\xi}2)\cdot Pr(\omega_2)} %
$$ 

As $Pr(\omega)$ is the same for all the $\omega$s, we get 
$$Pr(\omega_1 \lor \omega_2 | {\xi}1 \land  {\xi}2) %
$$ $$ %
(Pr(\omega_1|{\xi}1)\cdot Pr(\omega_1|{\xi}2)   %
+Pr(\omega_2|{\xi}1)\cdot Pr(\omega_2|{\xi}2) ) \cdot \frac{Pr( {\xi}1) \cdot Pr({\xi}2)}{Pr( {\xi}1 \land  {\xi}2)\cdot Pr(\omega)} %
$$ 

We can easily check that this translates to: 
$$Pl_{{\xi}1\land{\xi}2}(\{\omega_1,\omega_2\})= %
$$ $$ %
max( %
Pl_{{\xi}1}(\omega_1) \cdot Pl_{{\xi}2}(\omega_1) %
+ (Pl_{{\xi}1}(\{\omega_1,\omega_2\}-Pl_{{\xi}1}(\omega_1)) %
\cdot (Pl_{{\xi}2}(\{\omega_1,\omega_2\}-Pl_{{\xi}2}(\omega_1)) %
$$ $$ %
, %
Pl_{{\xi}1}(\omega_1) %
\cdot (Pl_{{\xi}2}(\{\omega_1,\omega_2\}-Pl_{{\xi}2}(\omega_2)) %
+ (Pl_{{\xi}1}(\{\omega_1,\omega_2\}-Pl_{{\xi}1}(\omega_1)) %
 \cdot Pl_{{\xi}2}(\omega_2)
$$ $$ %
, %
Pl_{{\xi}1}(\omega_2) %
\cdot (Pl_{{\xi}2}(\{\omega_1,\omega_2\}-Pl_{{\xi}2}(\omega_1)) %
+ (Pl_{{\xi}1}(\{\omega_1,\omega_2\}-Pl_{{\xi}1}(\omega_2)) %
 \cdot Pl_{{\xi}2}(\omega_1)
$$ $$ %
, %
Pl_{{\xi}1}(\omega_2) \cdot Pl_{{\xi}2}(\omega_2) %
+ (Pl_{{\xi}1}(\{\omega_1,\omega_2\}-Pl_{{\xi}1}(\omega_2)) %
\cdot (Pl_{{\xi}2}(\{\omega_1,\omega_2\}-Pl_{{\xi}2}(\omega_2)) %
)\cdot c$$ where $c$ is the normalizing factor mentioned earlier.

These formulas easily generalize for subsets of $\Omega$ with higher cardinality.
The normalizing factor should be chosen in such a way that 
$Pl_{{\xi}1\land{\xi}2}(\Omega)=1$.

The generalization of $\oplus_{Pl}$ for frames of discourse with cardinality higher than 3 runs along the following lines.
To combine $Pl_1$ with $Pl_2$ we calculate:
\begin{itemize}
\item
   for each subset $X$ of $\Omega$ \\
    $Pl_{result}(X)=PL_1^{\downarrow*X} \otimes_V Pl_2{\downarrow*X}$;         
\end{itemize}

The operator $\downarrow*{X}$ does only a change of the domain of the $Pl$ function keeping the values of $Pl$ for each subset of $X$ and presuming that the discourse frame consists only of $X$.
In this way we get  unnormalized $Pl$s here, which are not normalized during this operation.

The operator $\otimes_V$, returning a numerical value,
attempts 
identify such combinations of mass assignments $m_a$ and $m_b$ to singleton sets that will not violate the constraints imposed by 
plausibility functions $Pl_1$ and $Pl_2$ resp.  and such that 
the sum $\sum_{X; X a singleton} m_a(X)\cdot m_b(X)$ is maximal. 

This is done by the operation of 
so-called   pushing down the plausibilities to singleton sets. 
Independently for $Pl_1$ and $Pl_2$ candidate $m_a$ and $m_b$ are obtaining via "pushing-down" recursively a singleton $\omega$ of $\Omega$. A candidate $m_a$ is obtained if all singletons are pushed down. Different candidates are obtained by different sequences of pushing down. It is easy to imagine that the process is time-consuming and its complexity grows exponentially with the number of elements of a set. Nonetheless for small domains the operation is feasible.

The idea of the push-down operator $\downarrow+$ is as follows:
Let $Pl$ be a plausibility function. If $A$ does not contain $\omega$,  
$Pl^{\downarrow+\omega}(A)=min(Pl(A),Pl(A\cup\{\omega\})-Pl(\{\omega\}))$, and otherwise $Pl^{\downarrow+\omega}(A)=Pl(A)$. 

Under these conditions it is obvious that we do not seek actually the maximum product over the whole domain, but rather in some "corner points". We will give a formal proof elsewhere that this check is in fact sufficient to establish the maximum. Here we only want to draw attention to the analogy with linear programming, where we seek the maximum subject to linear constraints. Whenever we fix "pushdown" of one of the plausibility distributions, we in fact have a linear optimization case with the other. If found, we can do the same with the other. 

The $\oplus_{Pl}$ operator is characterized by commutativity and associativity. 
The commutativity is easily seen because all the operations are in fact symmetrical with respect to left and right hand of the operators. 
The associativity is more difficult to grasp, and a formal proof will be subject of another publication.
Nonetheless we can give here brief common-sense guidelines how it can be established. 
We can essentially concentrate on the associative properties of the maximum operator. Starting with the expression of combination of all the three plausibility functions, we can show that we can equivalently denote the same optimization task when drawing behind braces the first or the third operand. 

In the next section we show some properties of the new operator compared with Dempster rule of combination for some illustrative examples. 

\section{Examples}

Let us consider the  bodies of evidence in the tables 
\ref{ma}, \ref{mb}, \ref{mc}. 

\begin{table}\caption{mass function for the body of evidence ${\xi}a$}\label{ma}\begin{center}\begin{tabular}{|r|l|}\hline  m value &  for the set of elements \\  
\hline 0.0&  \{ \} \\  
\hline 0.25&  \{ red \} \\  
\hline 0.25&  \{ blue \} \\  
\hline 0.25&  \{ red , blue \} \\  
\hline 0.25&  \{ green \} \\  
\hline 0.0&  \{ red , green \} \\  
\hline 0.0&  \{ blue , green \} \\  
\hline 0.0&  \{ red , blue , green \} \\ \hline  
\end{tabular}\end{center}\end{table}

\begin{table}\caption{mass function for the body of evidence ${\xi}b$}\label{mb}\begin{center}\begin{tabular}{|r|l|}\hline  m value &  for the set of elements \\  
\hline 0.0&  \{ \} \\  
\hline 0.2&  \{ red \} \\  
\hline 0.4&  \{ blue \} \\  
\hline 0.1&  \{ red , blue \} \\  
\hline 0.0&  \{ green \} \\  
\hline 0.0&  \{ red , green \} \\  
\hline 0.3&  \{ blue , green \} \\  
\hline 0.0&  \{ red , blue , green \} \\ \hline  
\end{tabular}\end{center}\end{table} 

\begin{table}\caption{mass function for the body of evidence ${\xi}c$}\label{mc}\begin{center}\begin{tabular}{|r|l|}\hline  m value &  for the set of elements \\  
\hline 0.0&  \{ \} \\  
\hline 0.0&  \{ red \} \\  
\hline 0.15&  \{ blue \} \\  
\hline 0.25&  \{ red , blue \} \\  
\hline 0.35&  \{ green \} \\  
\hline 0.25&  \{ red , green \} \\  
\hline 0.0&  \{ blue , green \} \\  
\hline 0.0&  \{ red , blue , green \} \\ \hline  
\end{tabular}\end{center}\end{table} 

We can check the commutativity and obtain the results as in the table 
\ref{mab}.

\begin{table}\caption{mass function for the body of evidence $({\xi}a\oplus_{PL}{\xi}b)$
=$({\xi}b\oplus_{PL}{\xi}a)$
}\label{mab}\begin{center}\begin{tabular}{|r|l|}\hline  m value &  for the set of elements \\  
\hline 0.0&  \{ \} \\  
\hline 0.11111111111111105&  \{ red \} \\  
\hline 0.4999999999999999&  \{ blue \} \\  
\hline 0.22222222222222232&  \{ red , blue \} \\  
\hline 0.0&  \{ green \} \\  
\hline 0.0&  \{ red , green \} \\  
\hline 0.16666666666666674&  \{ blue , green \} \\  
\hline 0.0&  \{ red , blue , green \} \\ \hline  
\end{tabular}\end{center}\end{table} 

The associativity has been verified in table 
\ref{mabc}. 

\begin{table}\caption{mass function for the body of evidence $(({\xi}a\oplus_{PL}{\xi}b)\oplus_{PL}{\xi}c)$
=$({\xi}a\oplus_{PL}({\xi}b\oplus_{PL}{\xi}c))$
}\label{mabc}\begin{center}\begin{tabular}{|r|l|}\hline  m value &  for the set of elements \\  
\hline 0.0&  \{ \} \\  
\hline 0.0&  \{ red \} \\  
\hline 0.4214285714285715&  \{ blue \} \\  
\hline 0.3214285714285714&  \{ red , blue \} \\  
\hline 0.014285714285714124&  \{ green \} \\  
\hline 0.07142857142857151&  \{ red , green \} \\  
\hline 0.1357142857142858&  \{ blue , green \} \\  
\hline 0.0357142857142857&  \{ red , blue , green \} \\ \hline  
\end{tabular}\end{center}\end{table} 

It is worth noting, that the new operator is different from Dempster rule, compare tables \ref{mab} and \ref{mabDS}

\begin{table}\caption{mass function for the body of evidence $({\xi}a\oplus {\xi}b)$}\label{mabDS}\begin{center}\begin{tabular}{|r|l|}\hline  m value &  for the set of elements \\  
\hline 0.0&  \{ \} \\  
\hline 0.20833333333333337&  \{ red \} \\  
\hline 0.6249999999999999&  \{ blue \} \\  
\hline 0.04166666666666663&  \{ red , blue \} \\  
\hline 0.12500000000000003&  \{ green \} \\  
\hline 0&  \{ red , green \} \\  
\hline 0&  \{ blue , green \} \\  
\hline 0&  \{ red , blue , green \} \\ \hline  
\end{tabular}\end{center}\end{table} 

With this and other experiments we see clearly the tendency of Dempster rule to move mass downwards to singleton sets, whereas the new rule is much more cautious here and in fact does not introduce the feeling of certainty where it is not justified.

\section{Conclusions}
We have introduced in this paper a new DST operator for combining independent evidence providing a clear probabilistic definition of the plausibility function, which is preserved under this rule of 
combination.

We have also provided several toy examples to give an impression what results are returned by the new operator.

Though the strict theoretical proof of properties like cummutativeness, associativeness is still to be provided, the computations for test examples show that the properties really hold. 
It is also obvious from the examples that the new rule differs from the Dempster rule of evidence combination. 
An interested reader is invited to visit the Web page \linebreak
{\it http://www.ipipan.waw.pl/\~{}klopotek/DSTnew/DSTdemo.html}
to try out himself.

\end{document}